\documentclass{article}
\usepackage{spconf,amsmath,graphicx}
\usepackage{stfloats}
\usepackage{multirow}
\usepackage{bbding}

\title{Skeleton-based action analysis for ADHD diagnosis}
\name{Yichun Li$^{1}$, Yi Li$^2$, Rajesh Nair$^3$, Syed Mohsen Naqvi$^1$}
\address{\textsuperscript{\rm 1}{ Intelligent Sensing and Communications Research Group, Newcastle University, UK}\\
\textsuperscript{\rm 2}{School of Computing and Communications, Lancaster University, UK}\\
\textsuperscript{\rm 3}{ Cumbria, Northumberland, Tyne and Wear (CNTW), NHS Foundation Trust, UK}}
\begin{document}
%\ninept
%
\maketitle
\begin{abstract}
Attention Deficit Hyperactivity Disorder (ADHD) is a common neurobehavioral disorder worldwide. While extensive research has focused on machine learning methods for ADHD diagnosis, most research relies on high-cost equipment, e.g., MRI machine and EEG patch. Therefore, low-cost diagnostic methods based on the action and behavior characteristics of ADHD are desired. Skeleton-based action recognition has gained attention due to the action-focused nature and robustness. In this work, we propose a novel ADHD diagnosis system with a skeleton-based action recognition framework, utilizing a real multi-modal ADHD dataset and state-of-the-art detection algorithms. Compared to conventional methods, the proposed method shows cost-efficiency and significant performance improvement, making it more accessible for a broad range of initial ADHD diagnoses. Through the experiment results, the proposed method outperforms the conventional methods in accuracy and AUC. Meanwhile, our method is widely applicable for mass screening.
\end{abstract}

\section{Introduction}
\label{sec:intro}

Attention Deficit Hyperactivity Disorder (ADHD) is a common neurobehavioral and neurodevelopmental disorder affecting 2-5\% of school-age children worldwide, with a high rate of undiagnosed cases among adults \cite{nash2022machine}. %Conventional ADHD diagnosis is time-consuming and labor-intensive, requiring historical data from psychiatrists and pediatricians\cite{loh2022automated}. 
Recently, machine learning methods based on Magnetic Resonance Imaging (MRI) \cite{tang2022adhd2} and Electroencephalography (EEG) \cite{tang2020high2} have achieved high accuracy of over 95\% on related datasets \cite{khan2021novel}, i.e., ADHD-200, but are limited by their expensive equipment and high operational costs \cite{loh2022automated}. Thus, there is a need for machine learning methods based on low-cost data categories, e.g., video and audio, to facilitate ADHD screening and primary diagnoses.

According to the Diagnostic and Statistical Manual of Mental Disorders, Fifth Edition (DSM-V), supportive evidence shows that ADHD behavioral features such as fidgeting and restlessness in clinical notes are typically generalized rather than characterized by repetitive, stereotyped movements \cite{edition2013diagnostic}. Conventional clinical observation of ADHD behavioral features is limited by the difficulty in accurately counting and extracting these characteristics \cite{khan2021novel,dubreuil2020deep}. Recently, action recognition methods have overcome this limitation by extracting human skeleton-joint information from videos, removing irrelevant information, and being robust to dynamic environments and complex backgrounds \cite{duan2022revisiting}. In this paper, a novel action recognition method based on the human skeleton-joint modality toward ADHD diagnosis by identifying and analyzing raw video recordings. Our main contributions include: 1) designing and implementing a test focusing on ADHD actions and reaction ability, recorded through three cameras; 2) implementing and evaluating a novel ADHD diagnosis system based on action recognition networks; 3) proposing classification criteria to provide diagnosis results and analysis of ADHD behavioral characteristics; 4) verifying the efficiency and feasibility of the system and 5) reporting the whole process data and results to CNTW-NHS Foundation Trust for review by medical consultants/professionals and public dissemination in due course.

\section{Proposed framework}
\label{sec:format}
%Fig.1 shows the recording room layout and the 3-viewed recorded videos for the ADHD dataset.
%\begin{figure*}[!ht]
%\centering
%\includegraphics[scale=0.8]{images/Fig3.png}
%\caption{Recording room layout for the intelligent sensing (audio-video-torch/keypad sensors) ADHD dataset. Camera 1 records the facial information and upper torso information. Camera 2 \& 3 Record information of left and right body and limbs.
%}
%\end{figure*}
\subsection{Participants and Procedure}

This study utilizes a real recorded multi-modal ADHD dataset consisting of 7 adult ADHD subjects diagnosed by medical consultants under DSM-V criteria and 10 neurotypical controls without ADHD diagnosis history. The dataset includes 3 males and 4 females with ADHD and 9 males and 1 female in the control group. All subjects are provided by the CNTW-NHS Foundation Trust, while the control group volunteers are from Newcastle University.

An attention and responsiveness test is provided for all participants. We prepare four continuous dialogue tasks: 1) a brief conversation between the participants and the interviewer, approximately 10-20 minutes long, consisting of 21 questions; 2) performing Cambridge Neuropsychological Test Automated Battery (CANTAB) tasks, including Cambridge Gambling Task (CGT), Stop Signal Task (SST), Rapid Visual Information Processing (RVP), and Spatial Working Memory (SWM). This task takes about 40-50 minutes; 3) beep reaction task. This task takes 6 minutes; 4) watching videos, including a math video labelled `boring' and a rally video labelled `exciting'. This task takes 10 minutes.
% \begin{figure}[h!]
% \centering
% \vspace{-0.9em}
% \includegraphics[scale=0.32]{images/Test.png}
% \caption{ The proposed multi-modal ADHD real dataset. }
% \end{figure}

Videos are recorded by 3 GoPro cameras which contain a front-faced Camera 1 to record facial information. Meanwhile, Camera 2\&3 records information on the left and right sides of torso and limbs, respectively. The resolution of the three cameras is $3840\times 2160$.
\vspace{-0.5em}
\subsection{Proposed ADHD Diagnosis System}

We propose an action-based ADHD diagnosis and analysis system, which can be used in ADHD diagnosis with raw RGB videos and be a competitive approach to clinical, EEG, and FMRI diagnosis approaches. As aforementioned, compared to conventional fMRI and EEG-based methods, the proposed method is simple yet efficient because the video signal is easy to obtain with the low equipment cost. %Our proposed method achieves high accuracy with improved robustness, interoperability, and scalability.
The framework overview is shown in Fig. 1. Details of each task will be covered in the following sections.

\begin{figure}[!ht]
\centering
\vspace{-0.5em}
\includegraphics[scale=0.56]{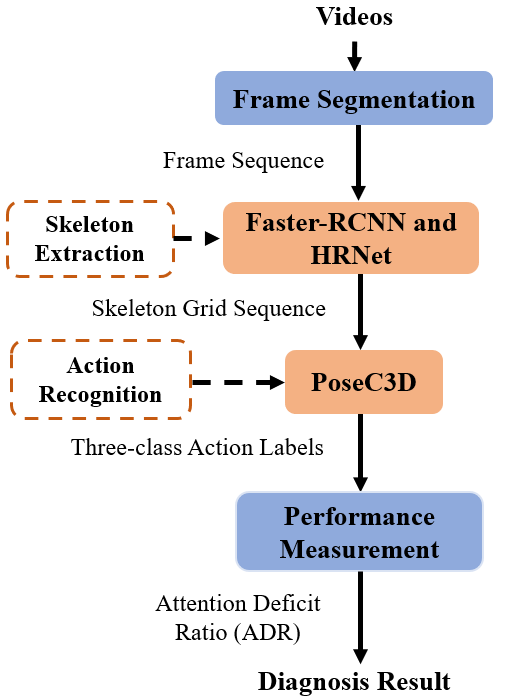}
\caption{Flow diagram of the ADHD diagnosis system. The dashed lines are the format of each tasks, while the solid lines point to the network and tasks of the system.}
\end{figure}

\subsection{Skeleton Extraction and Action Recognition}

In the frame segmentation task, the input video signal from our ADHD multi-modal dataset is decomposed into a frame sequence of 25 FPS. We use the \emph{detector} and \emph{estimator} to capture pose information in the frame sequence and record it as a human skeleton-joint grid sequence in the skeleton extraction task. %In skeleton extraction and action recognition task, pose information includes skeleton-joint sequences extracted from raw RGB video signals. 
Through these tasks, action-related information is extracted without contextual nuisances, such as background variation and unrelated personnel interference \cite{sabater2021one}.

In general, 2D poses are of better quality and higher accuracy than 3D poses \cite{liu2019skepxels}, which is crucial for applications related to medical diagnosis. In this work, a ResNet50-based Faster-RCNN network is used as the \emph{detector} \cite{sun2019deep}; the pose \emph{estimator} is a pre-trained HRNet because they achieve the state-of-the-art results on n the MPII (top) and COCO (bottom) datasets \cite{sun2019deep}. As shown in Fig. 2, we use Top-Down pose estimators to capture standard benchmarks such as COCO-keypoints of subjects and controls in a sitting position\cite{duan2022revisiting}. 17 joint points are detected and used in pose and action detection. Skeleton-joint grid sequence information is stored in a series of coordinate triplets $(x,y,c)$, where $c$ is related to the number of joints, height, and weight at each frame. $(x,y)$ is the corresponding coordinate of the $c$ \cite{duan2022revisiting}.

\begin{figure}[!ht]
\centering
\vspace{-0.5em}
\includegraphics[scale=0.47]{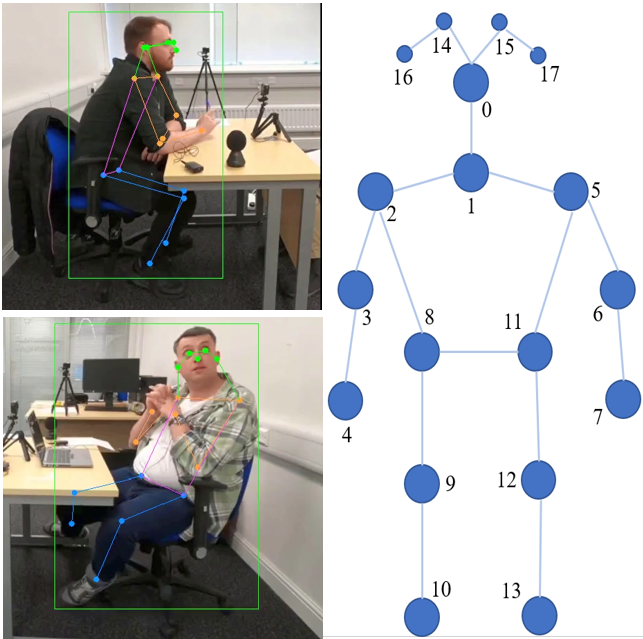}
\caption{ Skeleton extraction results from left and right side cameras and the COCO-Keypoints information. }
\end{figure}
\vspace{-0.5em}
\subsection{Performance Measurement}

Focus on the ADHD typical symptoms, the actions of subjects and controls in our dataset mainly contain three categories: still position, small ranges of limb fidgets, and large rotations of torso movements. The existing action recognition evaluation criterion cannot be applied to ADHD-specific classification and evaluate its action frequency characteristics. We propose a novel Hyperactivity Score ($HS$) and a measurement named Attention Deficit Ratio ($ADR$) as the evaluation criterion for action classification of ADHD symptoms detection. They focuses on the action change frequency of the subjects and controls during the test, which are also defined as model's ability to focus on the movement or posture. The Hyperactivity Score ($HS$) is calculated as :

\begin{equation}
HS_{n} = \begin{cases} HS_{n-1}+1 , & \text { if } l_{n}=l_{n-1} \\ HS_{n-1}-1, & \text { otherwise }\end{cases}
\end{equation}
where $n$ denotes the number of labels, $HS_{n}$ denotes the score of $n$ labels, $l_{n}$ denotes the $n$th label in the label sequence. $HS$ is rewarded if the action is consistent in the continuous time frame. Otherwise, it is punished.

\begin{table*}[h!]
\caption{The Attention Deficit Ratio ($ADR$) comparisons for the overall subjects and controls. And 'S', 'C', 'F', 'M' indicate subject, controls, female, and male, respectively. Each result is the average of 5 experiments.\\}
\centering
\small
\begin{tabular}{c|c c c c c c c c c c}
\hline
 \multirow{2}*{\textbf{Samples}}&S2 (M)&S6 (M)&S9 (F)&S10 (F)&S12 (F)&S13 (M)&S14 (F) & & &\\&C1 (M)&C3 (M)&C4 (M)&C5 (M)&C7 (M)&C8 (M)&C11 (M)&C15 (M)&C16 (F)&C17 (M)\\
%\hline
% \multirow{2}*{\textbf{ADR_{R}}}  & 0.43 & 0.76 & 0.74 & 0.73 & 0.70 & 0.70 & 0.79 & & &\\& 0.80 & 0.82 & 0.81 & 0.86 & 0.70 & 0.77 & 0.79 & 0.87 & 0.64 & 0.83 \\ 
 %\hline
 %\multirow{2}*{\textbf{ADR_{L}}}  & 0.54 & 0.75 & 0.75 & 0.71 & 0.75 & 0.74 & 0.72 & & &\\& 0.82 & 0.77 & 0.87 & 0.77 & 0.65 & 0.91 & 0.83 & 0.86 & 0.67 & 0.90 \\ 
\hline
\multirow{2}*{\textbf{ADR$_{Avg}$($\%$)}}  & 48.9 & 75.7 & 74.6 & 71.9 & 72.6 & 72.1 & 76.1 & & &\\& 80.8 & 79.9 & 83.9 & 81.7 & 67.8 & 84.0 & 81.3 & 86.6 & 65.6 & 86.5 \\
\hline\end{tabular}
\end{table*}

According to the effect of video length on $HS$, we normalize the results by the ratio of $HS$ to the $n$ labels and denote it as $ADR$, which is calculated as:

\begin{equation}
ADR(\%)=\frac{HS_{n}\cdot100 }{n} 
\end{equation}

%where $n$ denotes the number of labels, $HS_{n}$ denotes the score of $n$ period labels.
We use $ADR_{L}$ and $ADR_{R}$, i.e., $ADR$ measures of left and right viewpoints recording for two cameras, respectively. The final $ADR$ is the average $ADR_{L}$ and $ADR_{R}$.
%We use Camera 2 and Camera 3 for left and right viewpoints, respectively. Therefore, the average ADR measurement of $ADR_{L}$ and $ADR_{R}$, i.e., ADR measures of left and right viewpoint recording cameras, is calculated as:

%\begin{equation}
 %ADR_{Avg}=(ADR_{R}+ADR_{L})/2\\\\
%\end{equation}

%According to the ADR threshold, the system outputs 0 (predicted as ADHD subject) and 1 (predicted as control). 
The diagnosis results $R$ are obtained by binary classification of the $ADR$ results of all participants using a determined threshold $T$. The diagnosis result is calculated as :

\begin{equation}
R=\begin{cases} \text{ADHD}, & \text { if } ADR < T \\ \text{Control}, & \text { otherwise }\end{cases}
\end{equation}

The performance of the proposed ADHD diagnosis system is evaluated by the standard measurements, e.g., accuracy, sensitivity, precision, and the area under curve (AUC). 
%In classification results, the true positive rate (TPR) represents the ratio of \textit{positive} the classifier considers to all positive examples, false positive rate (FPR) represents the ratio of all \textit{negative} examples considered by the classifier to be positive. The area under the receiver operating characteristic curve is the AUC \cite{angelini20192d,lecun2015deep}.

\section{Experiments}
\label{sec:pagestyle}

\subsection{Dataset Preparation}
%\begin{figure}[!ht]
%\centering
%\includegraphics[scale=0.6]{images/Dataset.png}
%\caption{\footnotesize The real ADHD dataset recorded by Camera 2 (left) and Camera 3 (right) and the skeleton-joint based extracted posture }
%\end{figure}

We use a real multi-modal ADHD diagnosis dataset for the proposed ADHD diagnosis system. Especially to recognize ADHD symptom-related actions, a three-classes-action ADHD dataset is used for training and test in the action recognition.

The ADHD diagnosis dataset contains the left and right body information recorded by two side cameras. The whole dataset contains 34 videos. In the action recognition part, we divide the subjects' actions in the sitting state into three categories, i.e., still-position (Action 1), which contains 88 video clips, limb-fidgets (Action 2) with 110 clips, and torso movements (Action 3) with 101 clips. Each of the clips is between 10-15 seconds. The training, validation, and testing data split is 7/1/3, respectively.

The input frame is reduced from $3840\times 2160$ to $1080\times 920$ and down-sampled from 32 to 25 FPS to minimize the computation cost. 2D-Poses are captured and estimated by the top-down estimator from RGB inputs, as shown in Fig. 2. Actions are labeled per 50 frames in the training and diagnosis steps.

\subsection{Experiment Set up}

%For ADHD diagnosis, the target actions, i.e., limb fidgeting and slight body rotation, are micro-actions that are difficult to recognize accurately. 
We exploit a 3D-CNN structure (PoseC3D) as the main core network \cite{duan2022revisiting}. Different from commonly used GCN methods in skeleton-based action recognition, PoseC3D is a novel backbone that takes the 2D-Poses as the heatmap stacks of skeleton joints rather than graph coordinates. On the temporal dimension, the heatmap sequence of different time steps consists of a 3D-dimension heatmap volume. PoseC3D is more robust to the upstream pose estimation and temporal actions due to the 3D structure of heatmap \cite{duan2022revisiting}.  Compared with grid-based GCN methods, the interoperability of PoseConv3D makes it easier to involve human skeletons in multi-modality and multi-modal fusion, potentially used in ADHD diagnosis. Meanwhile, the PoseC3D performs better on most existing action detection datasets, such as UCF101, NTURGB-D, FineGYM, etc. \cite{cheng2020skeleton}.

Different from the original implementation \cite{duan2022revisiting}, the first convolution layer of our PoseC3D network is changed to 17$\times$25$\times$56$\times$56 kernels with 1$\times$1 stride to fit the size of our input data format. The training epochs for the action classification are 30, and the learning rate is $4\times 10^{-3}$. All the experiments are run on a workstation with four Nvidia GTX 1080 GPUs and 16 GB of RAM.

 %To the best of our knowledge, it is the first ADHD diagnosis work on simple video-based ADHD data \cite{loh2022automated}. We compare the proposed method with the other two commonly used skeleton-based action recognition networks (ST-GCN, MsG3D) in our action recognition function. To ensure the fairness of the experiment, we use the same configurations, i.e., estimator, detector, learning rate, and evaluation matrix, for ST-GCN and MS-G3d as the proposed PoseC3D. Based on this basic framework, we adapt two popular 3D structure networks: C3D and R3D, to the action recognition function. Different from skeleton-based methods, these two networks both use the raw RGB frame sequence as input. The training epochs for the action classification are 80, and the learning rate is $1\times 10^{-9}$.
 
\subsection{Time-Action based Diagnosis Results and Comparisons}

According to DSM-V, some symptoms of hyperactivity-impulsivity are observable in ADHD adults, such as difficulty sitting still, fidgeting legs, tapping with a pen, etc., and those actions are not characterized by repetitive stereotypical movements \cite{edition2013diagnostic}. However, it is hard to manually record irregular, high-frequency, and small-range actions during the traditional diagnostic process. Through our system, the skeleton-based poses and actions of each participant are fully captured and visualized. Fig. 3 shows the action recognition results timeline bar chart from a randomly selected subject and control.

\begin{figure}[!ht]
\centering
\includegraphics[scale=0.34]{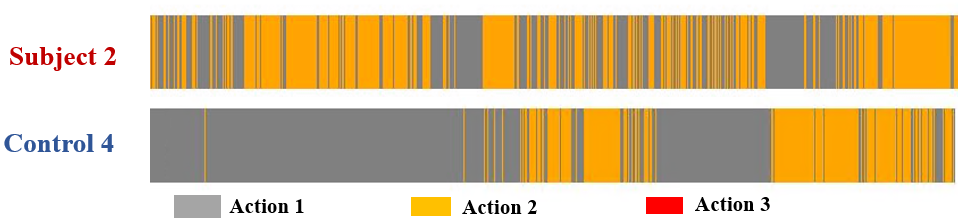}
\caption{ Action change timeline chart of a randomly selected subject and control. }
\end{figure}

Through Fig. 3, it can be easily observed that the action change frequency for the ADHD subject is significantly higher than the control. We further provide the $ADR$ performance of 7 subjects and 10 controls as shown in Table 1. 

% \begin{table*}[h!]
% \caption{\footnotesize The comparisons of the Attention Deficit Ratio (ADR) for the overall subjects and controls. And 'S', 'C', 'F', 'M' indicate subject, controls, female, and male, respectively. Each result is the average of 5 experiments.\\}
% \centering
% \scriptsize
% \begin{tabular}{c|c c c c c c c c c c}
% \hline
%  \multirow{2}*{\textbf{Samples}}&S2 (M)&S6 (M)&S9 (F)&S10 (F)&S12 (F)&S13 (M)&S14 (F) & & &\\&C1 (M)&C3 (M)&C4 (M)&C5 (M)&C7 (M)&C8 (M)&C11 (M)&C15 (M)&C16 (F)&C17 (M)\\
% %\hline
% % \multirow{2}*{\textbf{ADR_{R}}}  & 0.43 & 0.76 & 0.74 & 0.73 & 0.70 & 0.70 & 0.79 & & &\\& 0.80 & 0.82 & 0.81 & 0.86 & 0.70 & 0.77 & 0.79 & 0.87 & 0.64 & 0.83 \\ 
%  %\hline
%  %\multirow{2}*{\textbf{ADR_{L}}}  & 0.54 & 0.75 & 0.75 & 0.71 & 0.75 & 0.74 & 0.72 & & &\\& 0.82 & 0.77 & 0.87 & 0.77 & 0.65 & 0.91 & 0.83 & 0.86 & 0.67 & 0.90 \\ 
%  \hline
%  \multirow{2}*{\textbf{ADR_{Avg}(\%)}}  & 48.9 & 75.7 & 74.6 & 71.9 & 72.6 & 72.1 & 76.1 & & &\\& 80.8 & 79.9 & 83.9 & 81.7 & 67.8 & 84.0 & 81.3 & 86.6 & 65.6 & 86.5 \\
%  \hline
% \end{tabular}
% \end{table*}

From Table 1, the average $ADR_{Avg}$ for 7 subjects and 10 controls are 71.7\% and 79.8\%, respectively. The average $ADR_{Avg}$ of all 17 participants is 76.5\%. Therefore, 76.5\% is adapted as the threshold for the ADHD diagnosis. 

%To the best of our knowledge, it is the first ADHD diagnosis work on simple video-based ADHD data \cite{loh2022automated}. 
In the next experiment, we compare the proposed method with the other two commonly used skeleton-based action recognition networks (ST-GCN, MS-G3D) in our action recognition task. To ensure the fairness of the experiment, we use the same configurations, i.e., estimator, detector, learning rate, and evaluation matrix, for ST-GCN and MS-G3D as the proposed PoseC3D. Based on this basic framework, we adapt two popular 3D structure networks,i.e., C3D and R3D, to the action recognition task. Different from skeleton-based methods, these two networks both use the raw RGB frame sequence as input. The training epochs for the action classification are 80, and the learning rate is empirically set to $1\times 10^{-9}$. %In classification results, the true positive rate (TPR) represents the ratio of \textit{positive} the classifier considers to all positive examples, false positive rate (FPR) represents the ratio of all \textit{negative} examples considered by the classifier to be positive. The area under the receiver operating characteristic curve is the AUC \cite{angelini20192d,lecun2015deep}.
 
We further calculate the precision, sensitivity, accuracy, and AUC of four comparison networks: ST-GCN, MS-G3D, C3D, and R3D \cite{cheng2020skeleton,liu2020disentangling, tran2015learning, tran2018closer}, and our proposed PoseC3D framework in Table 2. 

\vspace{-0.5em}
\begin{table}[ht]
\caption{ ADHD diagnosis system performance with different neural networks. \\ }
\centering
\small\addtolength{\tabcolsep}{-1pt}
\begin{tabular}{c|c c c c}
\hline
\multicolumn{1}{l|}{} &Precision(\%)&F1(\%) &Accuracy(\%)&AUC \\ \hline
R3D \cite{tran2018closer} & 58.8 & 74.0 & 58.8& 0.50  \\ \hline
C3D \cite{tran2015learning}  & 85.7  & 70.6 & 70.6 & 0.71 \\ \hline
ST-GCN \cite{cheng2020skeleton} & 100.0  & 75.0 & 76.4 & 0.70  \\ \hline
MS-G3D \cite{liu2020disentangling} & 85.7  & 70.6 & 70.6 & 0.72  \\ \hline
\textbf{PoseC3D} & \textbf{100.0} & \textbf{88.9} &\textbf{88.2} & \textbf{0.83} \\ \hline
\end{tabular}
\end{table}

From Table 2, the proposed PoseC3D is significantly higher than the C3D, R3D, ST-GCN and MS-G3D in precision, accuracy, F1 Score and AUC. The Posec3d takes the advantage of combining the skeleton grid and heatmap, which leads to improved performance on action recognition task, indicates a clear differentiation between ADHD subjects and controls in diagnosis outcomes, as well as improved diagnosis accuracy.

%The performance of the action recognition module assumes paramount significance in the overall functioning of the diagnostic system. It is imperative to ensure the high accuracy and robustness of the action detection network, as it has a critical bearing on the diagnostic efficacy of the system.

\subsection{Ablation Study}

In the ablation study experiment, the original ADHD-3 dataset is shattered and labelled as ADHD subjects and controls. C3D-1 is a diagnostic discriminant networks trained on this binary ADHD classification dataset. Apart from the binary ADHD classification dataset, PoseC3D add the skeleton information extraction task. PoseC3D-2 and C3D-2 are action recognition networks with and without skeleton extraction task trained on the three-class action dataset, as mentioned in Section 3.1, respectively. It is highlighted that the action recognition task and the ADR task are closely related and cannot be separated. The AUC results are shown in Table 3:
\vspace{-0.9em}
\begin{table}[ht]
\caption{ Ablation study results with AUC. \\ }
\centering
\small\addtolength{\tabcolsep}{-1pt}
\small
%\scriptsize
\begin{tabular}{c|c|c}
\hline
Task& Network & AUC \\ \hline
- & C3D-1& 0.50 \\ \hline
Skeleton & PoseC3D-1 & 0.59 \\ \hline
Action+ADR & C3D-2 & 0.71 \\ \hline
\textbf{Skeleton + Action + ADR} & \textbf{PoseC3D-2} & \textbf{0.83} \\ \hline
\end{tabular}
\end{table}

According to the results of ablation study, firstly, the action recognition module plays a important role in the overall diagnostic system which significantly improves diagnostic accuracy by extracting and classifying action features. Secondly, the skeleton extraction task improves the detection accuracy on the basis of the action recognition task by its robustness for the impact of environmental interference. 
%According to the results of ablation study, our proposed framework significantly improves diagnostic accuracy by extracting and classifying action features. Comparing the results of Table 2 and the results of the Skeleton + Action + ADR method, the performance of the action recognition module plays a critical role in the overall diagnostic system. A high level of accuracy and robustness of the action recognition network is vital to achieving optimal diagnostic efficacy. Inaccuracies or inconsistencies in the output of the action recognition can significantly impact the subsequent stages of the diagnosis pipeline, leading to suboptimal outcomes.
%Therefore, it is necessary to devise innovative techniques to improve the performance of the action recognition module and enhance the diagnostic efficiency of the overall system. 

\section{Conclusion}
\label{sec:typestyle}
This paper proposed an ADHD diagnosis system based on a skeleton-joints modality action recognition framework. Multi-camera video data were recorded as training and test data in this work. A novel measure named ADR was proposed to evaluate the attention deficit performance of the action recognition results. The experimental results demonstrated that our system outperforms state-of-the-art methods regarding precision, accuracy, and AUC with high efficiency. %our system was less expensive and suitable for a broad range of initial ADHD diagnoses. 
Our systems are cost-effective and easily integrable into clinical practice. In future work, we plan to expand the dataset to cover a real-world patient distribution and record more multi-modal data such as EEG and fMRI for fusion and evaluation of related results. Furthermore, we will focus on the effectiveness of deep learning models, particularly those based on graph convolutional networks and spatial-temporal architectures, to achieve superior results in action recognition task, thereby enabling the development of more efficient diagnostic systems for various applications.

\vfill\pagebreak

%\section{REFERENCES}
%\label{sec:refs}

% References should be produced using the BibTeX program from suitable
% BibTeX files (here: strings, refs, manuals). The IEEEbib.bst bibliography
% style file from IEEE produces an unsorted bibliography list.
% -------------------------------------------------------------------------
\bibliographystyle{IEEEbib}
\bibliography{Template}

\end{document}